\RequirePackage{fix-cm}

\documentclass[twocolumn]{svjour3}       
\smartqed  
\usepackage{graphicx}
\usepackage[numbers,sort&compress]{natbib}
\usepackage{amssymb}
\usepackage[hyphens]{url}
\usepackage[hidelinks]{hyperref}
\hypersetup{breaklinks=true}
\usepackage[table,xcdraw]{xcolor}
\hypersetup{colorlinks = true,linkcolor = blue,anchorcolor =red,citecolor = blue, filecolor = red, urlcolor = blue,
            pdfauthor=author}
\usepackage{titlesec}
\usepackage{url}
\journalname{Arxiv}
\begin{document}

\title{Deepfake Detection and the Impact of Limited Computing Capabilities
}


\author{Paloma Cantero-Arjona,
    Alfonso S\'anchez-Maci\'an
}


\institute{P. Cantero-Arjona \and
           A. S\'anchez-Maci\'an \at
           Universidad Carlos III de Madrid \\
           Avda. Universidad 30, Leganés, Madrid, Spain \\
              Tel.: +34-916249433\\
              alfonsan@it.uc3m.es
}

\date{Received: date / Accepted: date}

\maketitle

\begin{abstract}
The rapid development of technologies and artificial intelligence makes deepfakes an increasingly sophisticated and challenging-to-identify technique. To ensure the accuracy of information and control misinformation and mass manipulation, it is of paramount importance to discover and develop artificial intelligence models that enable the generic detection of forged videos. This work aims to address the detection of deepfakes across various existing datasets in a scenario with limited computing resources. The goal is to analyze the applicability of different deep learning techniques under these restrictions and explore possible approaches to enhance their efficiency.
\keywords{deepfakes \and artificial intelligence \and deep learning \and convolutional networks \and transformers.}
\end{abstract}
\section{Introduction}
\label{intro}
Today, artificial intelligence is one of the most important and rapidly developing branches of computer science. Its growth is reflected daily in the number of tools and content made available to users, allowing people to generate artificial content very easily for various purposes.

A primary challenge associated with the widespread deployment of AI is the absence of usage limitations, resulting in multiple issues related to the creation of so-called deepfakes.

Deepfakes are multimedia files (videos, images, or audio) that have been manipulated using artificial intelligence techniques to make them appear real. These can be used for various purposes, but they are often used to harm the reputation of individuals, including, for example, creating fake explicit content involving celebrities or spreading false political messages that appear entirely real. This has led to situations involving blackmail, disinformation, and manipulation among the masses. The European Union Agency for Cybersecurity (ENISA) identified \cite{enisa} advanced disinformation campaigns, including the use of deepfakes, as the second top emerging cybersecurity threat for 2030.

For this reason, there is a need to detect deepfakes. Due to the rapid evolution of artificial intelligence techniques, this task becomes increasingly challenging every day, as all the flaws that allow detection are quickly addressed. Therefore, research and the development of techniques that can differentiate between real and manipulated content are crucial to minimize the generation of misinformation as much as possible.

The objective of this paper is to investigate the behavior of various artificial intelligence models in detecting deepfakes across different datasets within a computing-constrained environment. Additionally, the paper aims to explore potential fine-tuning approaches to improve their results.

The rest of the paper is structured as follows: First, background information is introduced, reviewing the main topics related to deepfake detection, datasets, and previous work in the application of AI to this task. Next, materials and methods are presented, followed by the experiments and the obtained results. Finally, conclusions are outlined.

\section{Background}
\label{related}

\subsection{Deepfakes and Detection}
Deepfakes are multimedia content that has been digitally or synthetically altered through the use of deep learning models\cite{dfdc}.

Due to the high accessibility of artificial intelligence, there has been a notable increase in the appearance of deepfakes. These can be used beneficially for society, but they can also be used in harmful ways to tarnish the image of individuals or spread false information\cite{harmful}, affecting the credibility of the media.

Deepfakes, which can be of audio, video, or a combination of both, are further categorized into multiple types. Since the focus of the work is on video deepfakes, the main types of video deepfakes\cite{Juefei_Xu_2022} are listed below:

\begin{itemize}
\item \textit{Face Swap}: This type, also known as Identity Swap, replaces the identity of the original image with the identity of a target image.
\item \textit{Expression Swap}: The Expression Swap deepfake type modifies the facial expressions of an original image to generate a new image with the desired expressions. Additionally, other attributes (head pose, lips, eyebrows) are also modified to make the video coherent.
\item \textit{Attribute Manipulation}: Attribute manipulation involves modifying the facial properties of a real face to generate a new image with those attributes.
\item \textit{Entire Face Synthesis}: Entire Face Synthesis generates a fake, non-existent face from a random vector and introduces it into an existing image.
\end{itemize}

Deepfake detection is a binary classification problem in which one must distinguish between real and manipulated videos or images. Models that detect deepfakes require large datasets so they can be trained, which limits the improvement of their performance. There are multiple types of features through which deepfakes can be detected. Next, some of these features, as well as the artificial intelligence techniques used for this purpose, will be mentioned:

\begin{itemize}
\item Spatial Detection: These detection methods utilize visible or invisible artifacts in the spatial domain. Traditional methods attempt to find disparities at the pixel level, such as patterns or noise. However, these methods are not very robust against simple manipulations. Other methods are based on deep neural networks \cite{tariq2020convolutional}, using spatial features to enhance effectiveness and generalization ability. Face preprocessing is also used to determine whether it is a deepfake or not, introducing videos into simple classifiers after preprocessing.

\item Frequency-Based Detection: Frequency-based detection methods use the frequency domain to discover differences between real and artificial images since generative adversarial networks often introduce invisible artifacts in that domain. For example, McCloskey and Albright\cite{mccloskey} investigated the generative model and found that pixels have a limitation in saturation, which allowed them to classify images based on pixel saturation. Other studies focus on analyzing the frequency domain to discover differences between images. The main drawback of these methods is that they are not robust against unseen generative adversarial network models. In the study conducted by Bai et al. \cite{bai} in 2020, FGPD-FA is proposed, which extracts three different types of features in the frequency domain to discriminate deepfakes.

\item Biological Sign-Based Detection: There are methods that classify deepfakes based on biological inconsistencies because, by introducing variations in the original images, some inconsistencies are generated that can be helpful in detection. These can be poorly generated facial attributes such as corneal reflection\cite{corneal}, eye blinking\cite{blink}, or head pose\cite{headposes}. More complex features can also be used, such as heart rate calculation\cite{heartbeat}.

\end{itemize}

\subsection{Deepfake Datasets}
Due to the high amount of data required to train deepfake detection models, it is important to be aware of the available datasets and their characteristics. The ones used in this study are presented next.

The Deepfake Detection Challenge Dataset  \cite{dfdc} is a dataset created in 2019 by Facebook for the Deepfake Detection Challenge (DFDC), an event on Kaggle created with the goal of accelerating the discovery of new ways to detect video deepfakes. The dataset contains more than 128,000 videos with face swaps using 8 different methods of facial modification.

FaceForensics++\cite{fforensics} is a dataset of 1,000 manipulated videos using four different techniques (Deepfakes, Face2Face, FaceSwap, and NeuralTextures). In addition to the videos, it also includes deepfake models that allow for data generation and augmentation.

UADFV\cite{UADFV} is a dataset developed by the University at Albany in New York. The main objective of its development was to attempt to detect deepfakes through physical signals such as facial expressions, lip movements, and blinking.

DeepfakeTIMIT\cite{TIMIT} is a dataset of videos in which faces have been altered using techniques based on generative adversarial networks, created from the original deepfake creation algorithm based on autoencoders. This dataset consists of a total of 320 videos, in two qualities, high (HQ) and low (LQ). The audio in these videos remains unchanged and unaltered.

\subsection{AI techniques and Deepfakes}
Different artificial intelligence techniques have been applied to the problem of detecting deepfakes. Deep learning is one of these approaches. It is based on neural networks with multiple hidden layers. The number of these layers depends on the complexity of the data used for the learning stages \cite{dl-2}.

The main deep learning options applied to this problem include Convolutional Neural Networks (CNNs)\cite{cnn}, Recurrent Neural Networks (RNNs)\cite{rnn}, Long Short-Term Memory (LSTM)\cite{lstm}, Variational Autoencoders (VAEs) \cite{vae} and Transformers\cite{transformers}.

CNNs are neural networks used with network topology data. They consist of an input layer, an output layer, and multiple hidden layers. However, the hidden layers read the information and perform a mathematical convolution on the input data. This operation is usually a matrix multiplication or dot product. Then, a non-linear activation function like ReLU (Rectified Linear Unit) is applied, along with a pooling function, which modifies the output by considering nearby outputs\cite{dl-book}.

RNNs allow learning from sequential data. Their structure, which connects each neuron to itself in the form of a cyclic directed graph, enables the network to store information from previous inputs, allowing earlier information to influence the current output.

LSTM neural networks are a type of network that allows for handling longer dependencies between data. They contain connections to previous nodes, enabling the learning of complete sequences of data. Their structure consists of an input gate, a forget gate, and an output gate. The input gate selects the values that enter the state, the forget gate applies a sigmoid activation function that decides which information to retain, and the output gate determines the information to transfer to the next cell.

Autoencoders are a type of neural network that attempts to copy the input information to the output while performing dimensionality reduction. They consist of two parts: an encoder that encodes the information, and a decoder that reconstructs the information to try to obtain the original data. Variational autoencoders are autoencoders that, instead of encoding the information as a single point, encode it as a distribution across a latent space.

Transformers are a type of neural network that contains a sequence-to-sequence architecture, which means it allows you to input a sequence of elements and transform them into another sequence. In this context, they resemble the previously described LSTM networks; however, they add an attention component. With this mechanism, the encoder determines which elements in a sequence are important by assigning them different weights. Subsequently, the decoder pays attention to the elements with higher weight. Another significant difference compared to other neural networks is that the inputs include positional encoding, so the model always knows the order of the introduced elements. These features allow transformers to be a model that addresses certain issues in neural networks, such as long-range dependencies, gradient vanishing, or parallel computation.

\subsection{Related work}
There are many works that try to provide efficient methods to detect deepfakes. Next, some of them are described.

In \cite{I3D}, three methods for detecting deepfakes were analyzed, with a focus on the use of 3D convolutional neural networks that consider both the image and the temporal sequence. The authors compared three types of these networks: I3D, 3D ResNet, and 3D ResNeXT, all adapted for deepfake detection. I3D uses RGB frames as input to the model and replaces the 2D convolutional layers of the Inception model with 3D convolutions for spatiotemporal modeling. 3D ResNet and 3D ResNeXT extend the 2D equivalents with a dimension to provide this spatiotemporal approach. Three experiments were conducted on the FaceForensics++ dataset, including the detection of various manipulation techniques, each technique individually, and combinations of manipulations. The results showed that 3D convolutional networks outperformed other image-based detection algorithms, with correct classification rates (CCR) of 83.86\%, 85.14\%, and 87.43\% for 3D ResNet, 3D ResNeXT, and I3D, respectively. However, a decrease in CCR was observed when detecting videos with manipulation techniques not seen during training.

An ensemble of 5 different transformer models was used in \cite{MEVER}. The classification result is the average of the results from these five models. They were trained using DFDC and evaluated on the FaceForensics++, CelebDF-V2\cite{Celeb_DF_cvpr20}, and WildDeepFake (WDF)\cite{WDF} datasets using balanced accuracy (BA) and area under the curve (AUC) metrics. Experimental results show that this method performs better on the CelebDF and WDF datasets, achieving 82.75\% and 84.94\% balanced accuracy, respectively. Additionally, the performance on different types of manipulation is compared, with the best performance observed in the case of Face Swap manipulation, achieving 78.40\%. Furthermore, the article examines adversarial robustness, revealing vulnerability to the Projected Gradient Descent (PGD) attack.

In \cite{ISTVT}, the authors proposed a spatiotemporal transformer model that incorporates self-attention and a mechanism for capturing spatial artifacts and temporal inconsistencies. They used a multi-layer perceptron for prediction and conducted experiments on datasets such as FaceForensics++, FaceShifter\cite{faceshifter}, DeeperForensics\cite{deeperforensics}, Celeb-DF, and DFDC. The results demonstrated that their approach outperformed convolutional neural network-based methods in most cases, achieving over 90\% accuracy, highlighting the potential of transformer-based methods.
They also assessed the model's generalization ability across different datasets using the area under the curve (AUC) metric, and it performed better than existing methods. However, the generalization capability decreased notably in the Celeb-DF and DFDC datasets due to the model's lack of knowledge about them.
Additionally, the model's robustness to perturbations like JPEG compression, scaling, and random dropout was evaluated, and it outperformed state-of-the-art methods, showcasing its superior performance and robustness.

Other techniques have been applied to this problem. LSTM was used in \cite{tariq}. By applying transfer learning techniques, they achieved up to 93.86\% accuracy on high-quality datasets. Another technique commonly used in the field is the use of various biological signals. The authors of \cite{rPPG} used an attention module in combination with Remote Photoplethysmography (rPPG), which uses light to determine heart rate and classify whether a video has been altered or not. This method achieves an average accuracy of 98.65\% on the FaceForensics++ dataset.

\section{Materials and methods}
\label{methods}
This works explores deepfake (face swap type) detection in a scenario with limited computation capabilities, using five different datasets: UADFV, DFTimit
LQ, DFTimit HQ, DFDC,and FaceForensics++.

To develop this work, various experiments have been conducted on a server with the Ubuntu Linux operating system. The server has the following configuration:
\begin{itemize}

    \item 4 CPUs - QEMU Virtual CPU version 2.5+ 64 bits
    \item RAM: 32 GB
    \item Storage memory: 2TB
\end{itemize}

The models have been implemented\footnote{https://github.com/PalomaCantero/Deepfake-Detection-and-the-Impact-of-Limited-Computing-Capabilities} in the Python programming language, and the main libraries used for this project have been Pandas, NumPy, PyTorch, OpenCV, Matplotlib, Scikit-learn, and einops. These libraries have allowed the use of information in multiple formats, as well as its transformation and processing in different artificial intelligence models or the graphical representation of the results of the experiments. To apply the selected models to the deepfake datasets, the workflow defined in \cite{otto2020dfperformancecomp} has been used as a foundation

The flow for data processing is as follows. First, the datasets are loaded and preprocessed. For this purpose, faces are extracted using Retina Face \cite{retina}, a Deep Learning model for face detection, using ResNet50 as the backbone network. This method, despite being developed in 2019, is currently one of the most effective methods for face detection \cite{wider}, evaluated on the WIDER Face dataset.

A specific number of frames is extracted from each video, which is determined through experimentation, and then faces are extracted from these frames. Horizontal or vertical flipping is applied to the extracted images to perform a small data augmentation.

Two detection models have been studied in this limited computational resources scenario. First, a 3DCNN model was implemented, more specifically, it was decided to use a (2+1)D CNN model \cite{3dVideo}. This neural network enables a decomposition in spatial and temporal dimensions, thereby increasing precision compared to conventional 3D convolutional models. The model uses (2+1)D convolutions with residual connections in the original ResNet18 model, replacing the original convolution with the mentioned (2+1)D convolution. The branch performs the computation, and gradients flow through the residual connections, avoiding the gradient vanishing problem.

Second, it was decided to use a transformer model since they are achieving the best results in the field. To initiate the detection system, the initial architecture was based on the ViViT model: A Video Vision Transformer\cite{vivit}. This architecture consists of a transformer designed for video classification. The first step is to encode the input videos using one of the available embeddings. As the model architecture, the "Factorised Encoder" Model 2 is chosen. This model consists of two encoders, one spatial that encodes interactions within the same temporal space, and another temporal that models interactions between tokens from different temporal spaces. Both the encoding and the model have been chosen considering the best results from the work and the available computational resources. Finally, a multi-layer perceptron is used as a classifier that uses a threshold value of 0.5 to determine the class to which the input video belongs.

\section{Experiments and Results}

In this section, several different experiments are described, oriented to adjust the parameters to find the best possible model. The loss function and the optimizer have been chosen beforehand based on the existing state-of-the-art solutions. As we face a binary classification problem, binary cross-entropy, which measures the distance between real and predicted classes, was chosen as the loss function. Adam optimizer was chosen as it has several advantages in methods including attention (e.g. transformers) such as its faster convergence due to the  sharpness reduction effect of adaptive coordinate-wise scaling \cite{pan2023toward}.

\subsection{Experiment 1: Number of Frames}
With the aim of finding the optimal number of frames for each dataset, experiments were conducted using pre-trained initial models with the following numbers: 5, 10, 15, 20, and 25. 

This experiment was carried out with all datasets except for DFDC, where the frame value was fixed at 5. This decision stems from the existing computational constraints and the substantial size of the DFDC dataset, rendering these experiments impractical for this specific case.

The experiments were conducted with the following configuration:
\begin{itemize}
    \item Learning rate: 0.0001.
    \item Episodes: 5.
    \item Batch Size: 32.
\end{itemize}

The pre-trained Xception model was used for the UADFV, DFTimit LQ, and DFTimit HQ datasets, and the EfficientNet B7 was used for the FaceForensics++ dataset. It should be noted that the good results in these tests are due to the fact that the model weights used were obtained by training with the same datasets. This is not a problem as these tests do not aim to find the best model but rather to determine the optimal number of frames to use.

Generally, it can be observed in tables \ref{frames-UADFV}, \ref{frames-LQ}, \ref{frames-hq}, and \ref{frames-FF} that the results are similar for these experiments at different numbers of frames, especially in precision. Therefore, to set the number of frames, the value of the loss function will be taken as a reference.

\begin{table}[ht!]
  \centering
\begin{tabular}{|c|c|c|}
\hline
\rowcolor[HTML]{C0C0C0} 
Frames & Precision  & Loss F.\\ \hline
\rowcolor[HTML]{FFFFFF} 
{\color[HTML]{1F2328} 5}  & {\color[HTML]{1F2328} 0.9748}          &  {\color[HTML]{1F2328} 0.0614}\\ \hline
\rowcolor[HTML]{EFEFEF} 10 & {\color[HTML]{1F2328} \textbf{0.9784}} & {\color[HTML]{1F2328} 0.0599}\\ \hline
\rowcolor[HTML]{FFFFFF} {\color[HTML]{1F2328} 15} & {\color[HTML]{1F2328} 0.9712}   & {\color[HTML]{1F2328} \textbf{0.0595}} \\ \hline
\rowcolor[HTML]{EFEFEF}
{\color[HTML]{1F2328} 20} & {\color[HTML]{1F2328} 0.9712}                  & {\color[HTML]{1F2328} 0.0618}\\ \hline
25 & \textbf{0.9784}  & 0.0695 \\ \hline
\end{tabular}
\caption{Results Experiment 1 - UADFV}
\label{frames-UADFV}
\end{table}

\begin{table}[ht!]
  \centering
\begin{tabular}{|c|c|c|}
\hline
\rowcolor[HTML]{C0C0C0} 
Frames & Precision & Loss F. \\ \hline
\rowcolor[HTML]{FFFFFF} 
{\color[HTML]{1F2328} 5} & {\color[HTML]{1F2328} 1.0} & {\color[HTML]{1F2328} 0.0012} \\ \hline
\rowcolor[HTML]{EFEFEF} 
10 & {\color[HTML]{1F2328} 1.0} & {\color[HTML]{1F2328} 0.0012} \\ \hline
\rowcolor[HTML]{FFFFFF} 
{\color[HTML]{1F2328} 15} & {\color[HTML]{1F2328} 1.0}  & {\color[HTML]{1F2328} \textbf{0.0011}} \\ \hline
\rowcolor[HTML]{EFEFEF} 
{\color[HTML]{1F2328} 20} & {\color[HTML]{1F2328} 1.0}  & {\color[HTML]{1F2328} 0.0012} \\ \hline
25 & \cellcolor[HTML]{FFFFFF}{\color[HTML]{1F2328} 1.0} & \cellcolor[HTML]{FFFFFF} 0.0024 \\ \hline
\end{tabular}
\caption{Results Experiment 1  - DFTimit LQ}
\label{frames-LQ}
\end{table}

\begin{table}[ht!]
  \centering
\begin{tabular}{|c|c|c|}
\hline
\rowcolor[HTML]{C0C0C0} 
Frames & Precision & Loss F. \\ \hline
\rowcolor[HTML]{FFFFFF} 
{\color[HTML]{1F2328} 5} & {\color[HTML]{1F2328} 1.0}  & {\color[HTML]{1F2328} 0.0019} \\ \hline
\rowcolor[HTML]{EFEFEF} 
10 & {\color[HTML]{1F2328} 1.0} & {\color[HTML]{1F2328} \textbf{0.0017}} \\ \hline
\rowcolor[HTML]{FFFFFF} 
{\color[HTML]{1F2328} 15} & {\color[HTML]{1F2328} 0.9996} &  {\color[HTML]{1F2328} 0.0035} \\ \hline
\rowcolor[HTML]{EFEFEF} 
{\color[HTML]{1F2328} 20} & {\color[HTML]{1F2328} 1.0} & {\color[HTML]{1F2328} 0.0023} \\ \hline
25 & \cellcolor[HTML]{FFFFFF}{\color[HTML]{1F2328} 1.0}  & 0.0018 \\ \hline
\end{tabular}
\caption{Results Experiment 1  - DFTimit HQ}
\label{frames-hq}

\end{table}

\begin{table}[ht!]
  \centering
\begin{tabular}{|c|c|c|}
\hline
\rowcolor[HTML]{C0C0C0} 
Frames & Precision & Loss F. \\ \hline
\rowcolor[HTML]{FFFFFF} 
{\color[HTML]{1F2328} 5} & {\color[HTML]{1F2328} 0.9893} & {\color[HTML]{1F2328} 0.0263} \\ \hline
\rowcolor[HTML]{EFEFEF} 
10 & {\color[HTML]{1F2328} 0.9890}  & {\color[HTML]{1F2328} 0.0260} \\ \hline
\rowcolor[HTML]{FFFFFF} 
{\color[HTML]{1F2328} 15} & {\color[HTML]{1F2328} \textbf{0.9899}}  & {\color[HTML]{1F2328} \textbf{0.0259}} \\ \hline
\rowcolor[HTML]{EFEFEF} 
{\color[HTML]{1F2328} 20} & {\color[HTML]{1F2328} 0.9874} & {\color[HTML]{1F2328} 0.0360} \\ \hline
25 & \cellcolor[HTML]{FFFFFF}{\color[HTML]{1F2328} 0.9799} & 0.0398 \\ \hline
\end{tabular}
\caption{Results Experiment 1  - FaceForensics++}
\label{frames-FF}

\end{table}

Considering that most of the results show 15 frames as the optimal parameter, except for experiment 3 with the DFTimit HQ dataset, it was decided to set the parameter for all datasets to 15 frames and 5 frames for DFDC.

\subsection{Experiment 2: 3DCNN model}
The first of the models used was the 3DCNN model described in section \ref{methods}. The objective of this experiment was to test convolutional networks applied to the detection of {deepfakes} to see if they were suitable for the use case with the specified limited capabilities.

Several experiments were conducted with the optimizer, loss function, and the number of {frames} previously described. In addition to splitting the data samples into training and validation sets, a subset of the data was also used for the final {test} of the model. The first experiment for this model was conducted with the following configuration:

\begin{itemize}
\item {Learning rate}: 0.0001.
\item {Episodes}: 10.
\item {Batch Size}: 32.
\end{itemize}

\begin{figure}[ht!]
	\centering
	\begin{minipage}{.45\columnwidth}
		\centering
		\includegraphics[width=\textwidth]{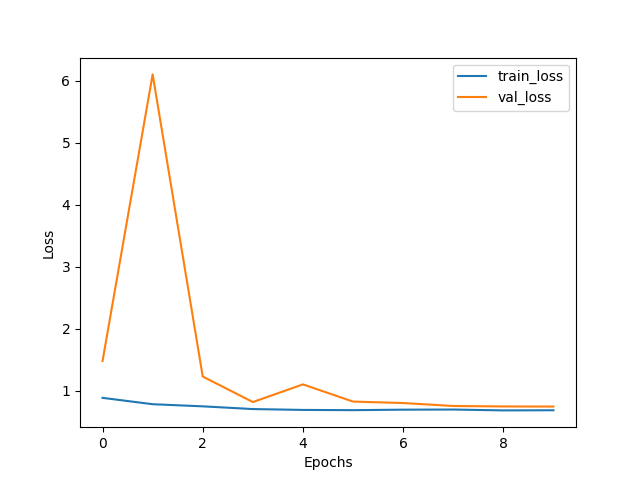}
		\caption{Exp. 3DCNN - UADFV}
		\label{expcnn3d.1}
	\end{minipage}%
	\begin{minipage}{.45\columnwidth}
		\centering
		\includegraphics[width=\textwidth]{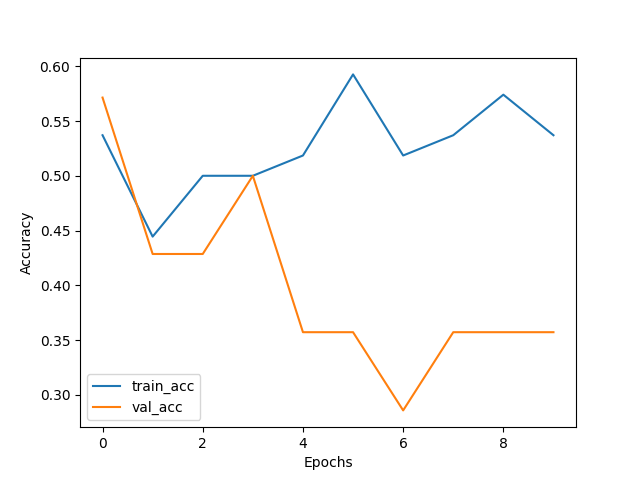}
		\caption{Exp. 3DCNN - UADFV}
		\label{expcnn3d.2}
	\end{minipage}
 	\begin{minipage}{.45\columnwidth}
		\centering
		\includegraphics[width=\textwidth]{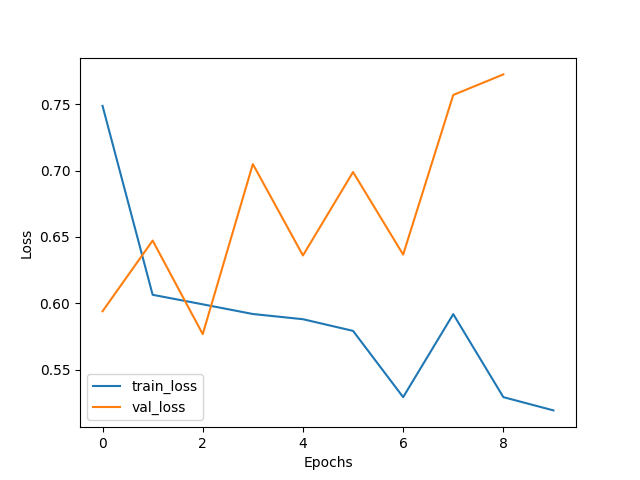}
		\caption{Exp. 3DCNN - DFTimit LQ}
		\label{expcnn3d.3}
	\end{minipage}
 	\begin{minipage}{.45\columnwidth}
		\centering
		\includegraphics[width=\textwidth]{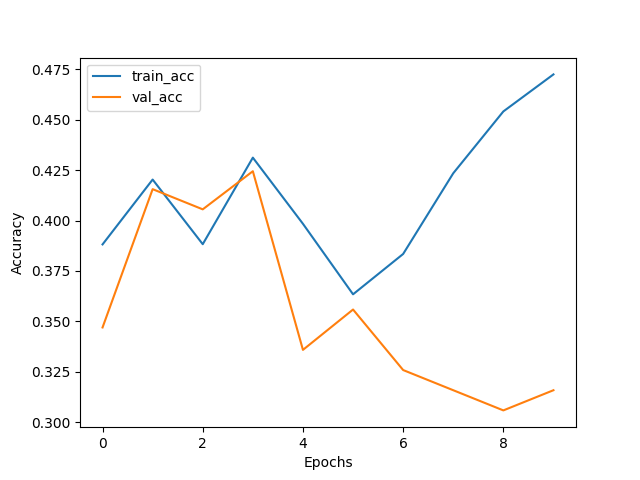}
		\caption{Exp. 3DCNN - DFTimit LQ}
		\label{expcnn3d.4}
	\end{minipage}
 	\begin{minipage}{.45\columnwidth}
		\centering
		\includegraphics[width=\textwidth]{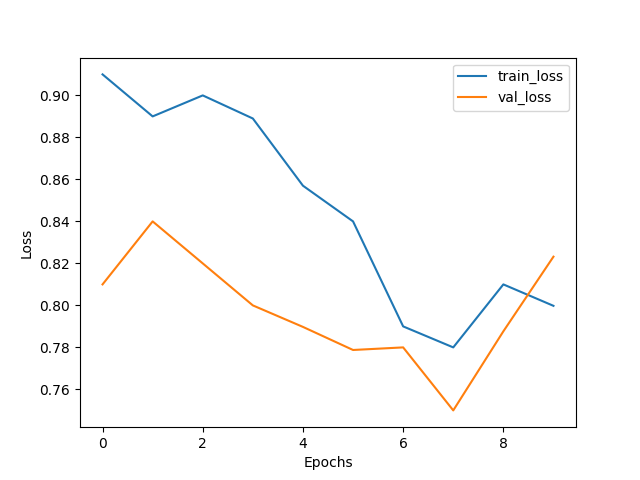}
		\caption{Exp. 3DCNN - DFTimit HQ}
		\label{expcnn3d.5}
	\end{minipage}
 	\begin{minipage}{.45\columnwidth}
		\centering
		\includegraphics[width=\textwidth]{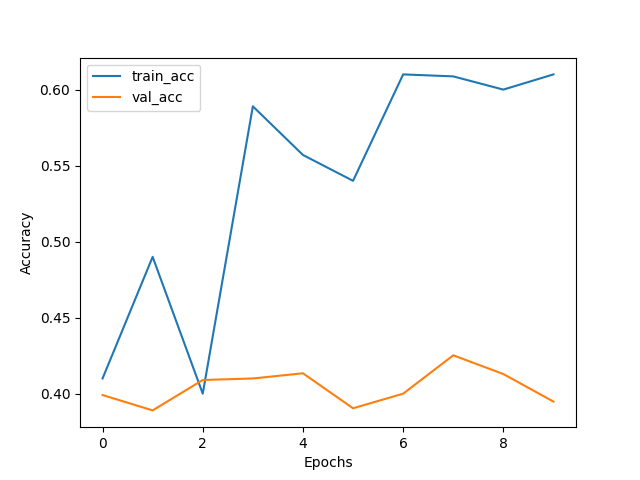}
		\caption{Exp. 3DCNN - DFTimit HQ}
		\label{expcnn3d.6}
	\end{minipage}
  	\begin{minipage}{.45\columnwidth}
		\centering
		\includegraphics[width=\textwidth]{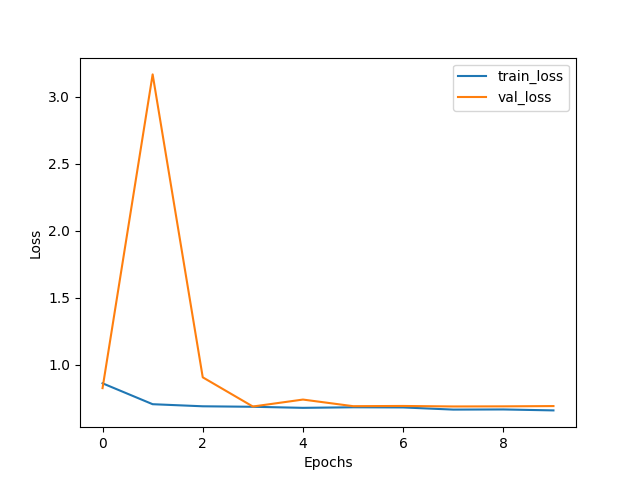}
		\caption{Exp. 3DCNN - DFDC}
		\label{expcnn3d.7}
	\end{minipage}
 	\begin{minipage}{.45\columnwidth}
		\centering
		\includegraphics[width=\textwidth]{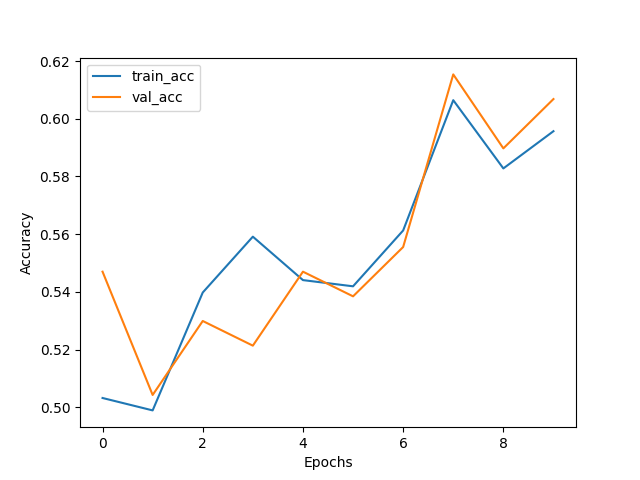}
		\caption{Exp. 3DCNN - DFDC}
		\label{expcnn3d.8}
	\end{minipage}
  	\begin{minipage}{.45\columnwidth}
		\centering
		\includegraphics[width=\textwidth]{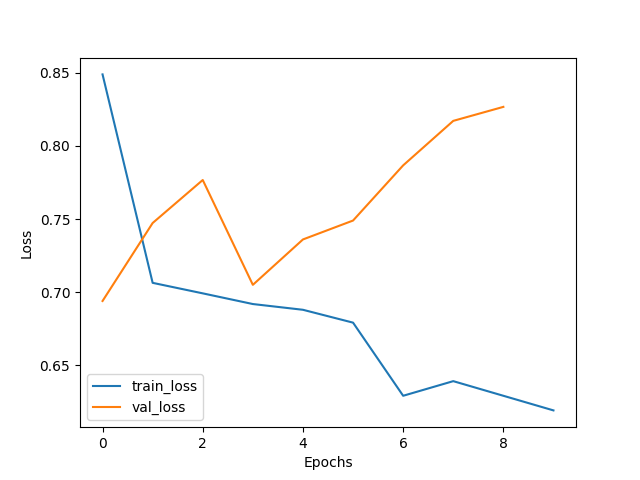}
		\caption{Exp. 3DCNN - FF++}
		\label{expcnn3d.9}
	\end{minipage}
 	\begin{minipage}{.45\columnwidth}
		\centering
		\includegraphics[width=\textwidth]{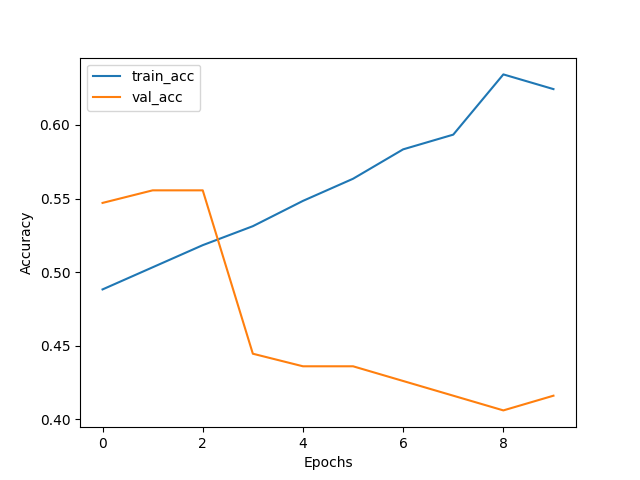}
		\caption{Exp. 3DCNN - FF++}
		\label{expcnn3d.10}
	\end{minipage}
\end{figure}

Figures \ref{expcnn3d.1}, \ref{expcnn3d.2}, \ref{expcnn3d.3}, \ref{expcnn3d.4}, \ref{expcnn3d.5}, \ref{expcnn3d.6}, \ref{expcnn3d.7}, \ref{expcnn3d.8}, \ref{expcnn3d.9}, and \ref{expcnn3d.10} show the learning curves of the executed models. Since the results did not indicate satisfactory outcomes, as the curves demonstrate that the model fails to learn completely or shows overfitting, it was decided to increase the learning rate to 0.001.

Again, upon observing the results, it is noted that the models struggle to adjust the frame characteristics to predict correctly. As the experimentation time with this model reached up to a total of 3454 minutes for the FaceForensics++ dataset, we determined that this model is highly susceptible to constraints on computational capabilities, rendering it an imprudent choice in this scenario.

Table \ref{2dcnn-exp} details the results of the experiments that yielded the best outcomes with the 3DCNN model.

\begin{table}
  \centering
\begin{tabular}{|c|c|c|c|}
\hline
\rowcolor[HTML]{C0C0C0} 
Dataset & Precision & Loss function  \\ \hline
\cellcolor[HTML]{FFFFFF}{\color[HTML]{1F2328} UADFV} & \cellcolor[HTML]{FFFFFF}{\color[HTML]{1F2328} 0.5250} & \cellcolor[HTML]{FFFFFF}{\color[HTML]{1F2328} 0.5780}  \\ \hline
\rowcolor[HTML]{EFEFEF} 
DFTimit LQ & {\color[HTML]{1F2328} 0.4732} & {\color[HTML]{1F2328} 0.6919}  \\ \hline
\cellcolor[HTML]{FFFFFF}{\color[HTML]{1F2328} DFTimit HQ} & \cellcolor[HTML]{FFFFFF}{\color[HTML]{1F2328} 0.4960} & \cellcolor[HTML]{FFFFFF}{\color[HTML]{1F2328} 0.7426} \\ \hline
\rowcolor[HTML]{EFEFEF} 
{\color[HTML]{1F2328} FaceForensics++} & {\color[HTML]{1F2328} 0.5830} & {\color[HTML]{1F2328} 0.6827} \\ \hline
DFDC & \cellcolor[HTML]{FFFFFF}{\color[HTML]{1F2328} 0.6279} & 0.9619 \\ \hline
\end{tabular}
\caption{Results Experiments 3DCNN}
\label{2dcnn-exp}
\end{table}

\subsection{Experiment 3: Vision Transformer model}

The first restriction found in this model due to the limited computation capabilities is related to encoding the input videos. Tubelet embedding was  initially proposed. This method extracts tokens that span the spatial-temporal domain, combining information from frames with temporal information. However, due to the high computational demands of this encoding, Patch embedding was used in order to get an appropriate batch size for experiments.

As previously pointed out in Section \ref{methods},  the ”Factorised Encoder” Model 2 was chosen among the ones proposed for ViViT as it was the one with best results taking into account the computational resources.
The initial configuration for the model was as follows:

\begin{itemize}
    \item Initial learning rate: 0.0001.
    \item Episodes: 100.
    \item Batch Size: 32.
    \item Transformer Dimension: 192.
    \item Transformer Depth: 4.
    \item Transformer Heads: 3.
    \item Transformer Head Dimension: 64.
\end{itemize}

\begin{figure}[ht!]
	\centering
	\begin{minipage}{.45\columnwidth}
		\centering
		\includegraphics[width=\textwidth]{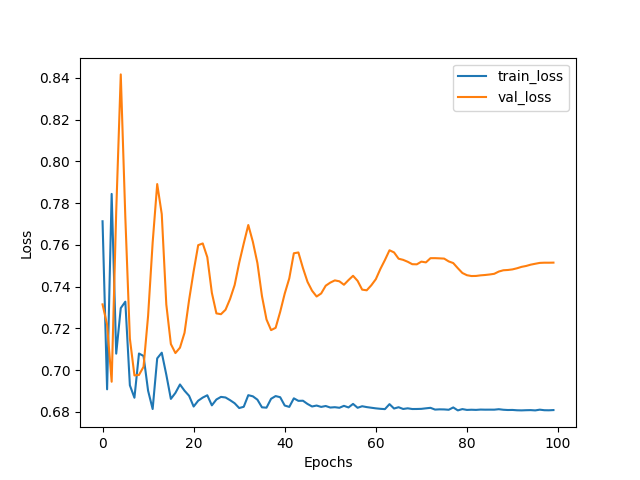}
		\caption{Exp. 1 ViT - UADFV}
		\label{exp1.1}
	\end{minipage}%
	\begin{minipage}{.45\columnwidth}
		\centering
		\includegraphics[width=\textwidth]{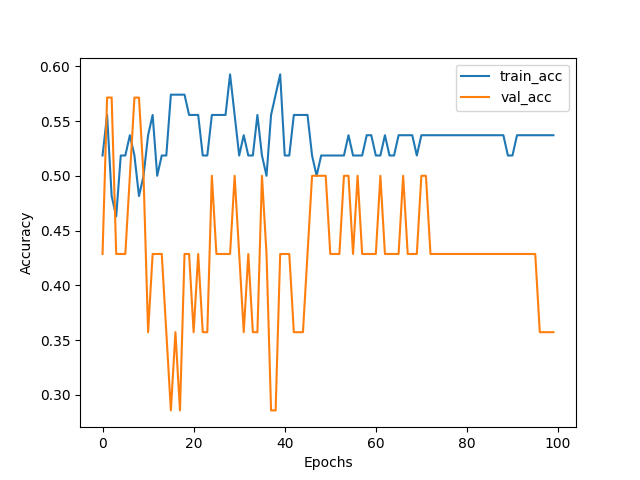}
		\caption{Exp. 1 ViT - UADFV}
		\label{exp1.2}
	\end{minipage}
 	\begin{minipage}{.45\columnwidth}
		\centering
		\includegraphics[width=\textwidth]{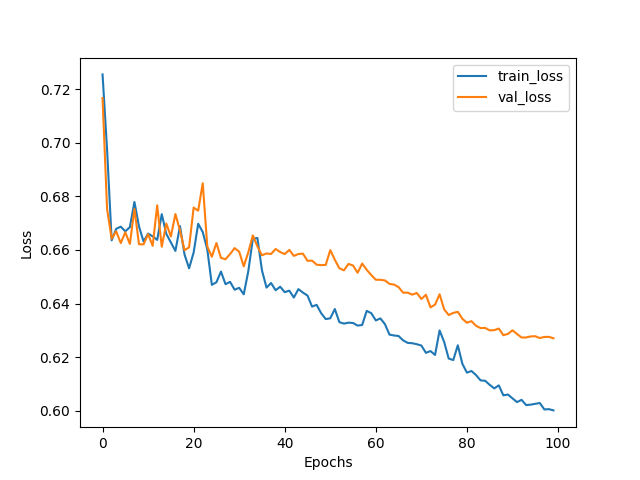}
		\caption{Exp. 1 ViT - DFTimit LQ}
		\label{exp1.3}
	\end{minipage}
 	\begin{minipage}{.45\columnwidth}
		\centering
		\includegraphics[width=\textwidth]{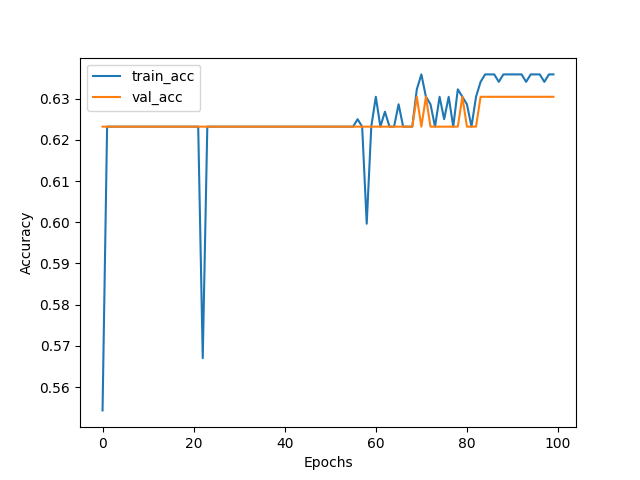}
		\caption{Exp. 1 ViT - DFTimit LQ}
		\label{exp1.4}
	\end{minipage}
 	\begin{minipage}{.45\columnwidth}
		\centering
		\includegraphics[width=\textwidth]{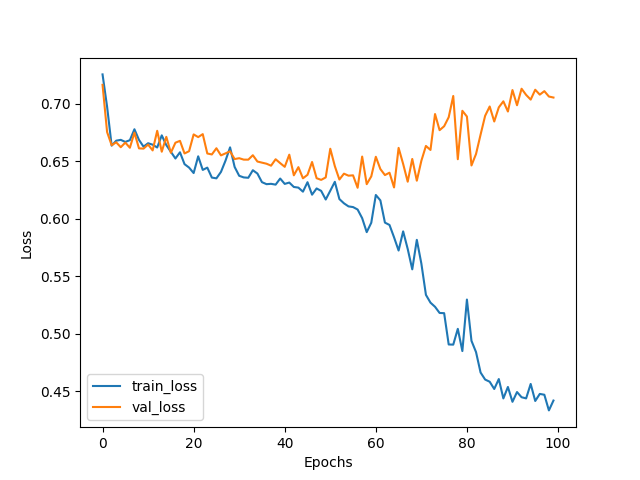}
		\caption{Exp. 1 ViT - DFTimit HQ}
		\label{exp1.5}
	\end{minipage}
 	\begin{minipage}{.45\columnwidth}
		\centering
		\includegraphics[width=\textwidth]{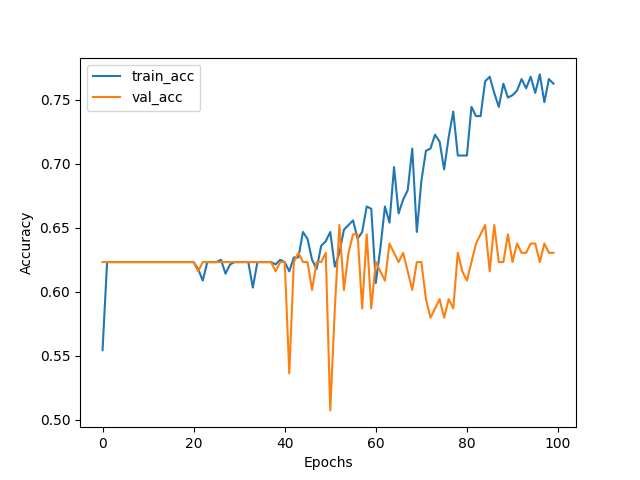}
		\caption{Exp. 1 ViT - DFTimit HQ}
		\label{exp1.6}
	\end{minipage}
  	\begin{minipage}{.45\columnwidth}
		\centering
		\includegraphics[width=\textwidth]{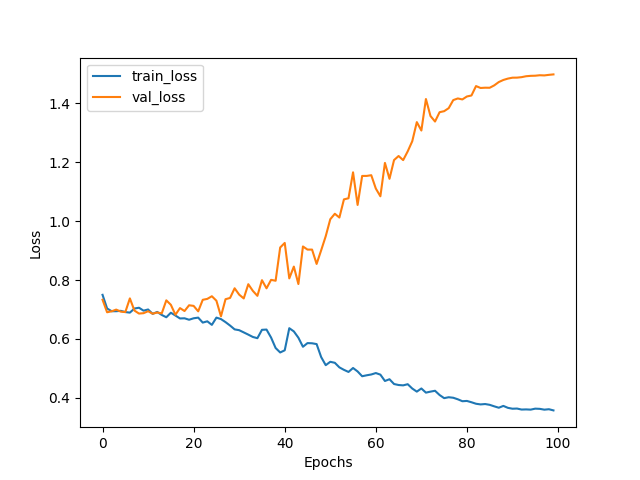}
		\caption{Exp. 1 ViT - DFDC}
		\label{exp1.7}
	\end{minipage}
 	\begin{minipage}{.45\columnwidth}
		\centering
		\includegraphics[width=\textwidth]{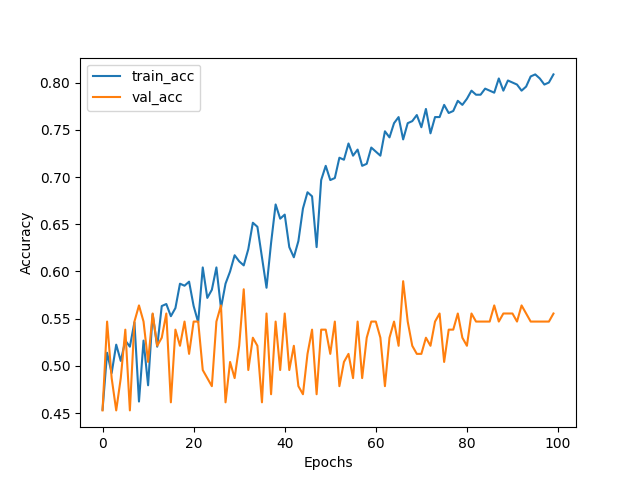}
		\caption{Exp. 1 ViT - DFDC}
		\label{exp1.8}
	\end{minipage}
  	\begin{minipage}{.45\columnwidth}
		\centering
		\includegraphics[width=\textwidth]{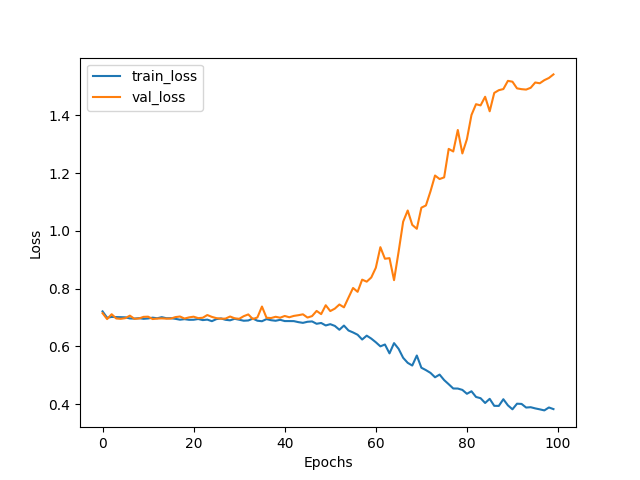}
		\caption{Exp. 1 ViT - FF++}
		\label{exp1.9}
	\end{minipage}
 	\begin{minipage}{.45\columnwidth}
		\centering
		\includegraphics[width=\textwidth]{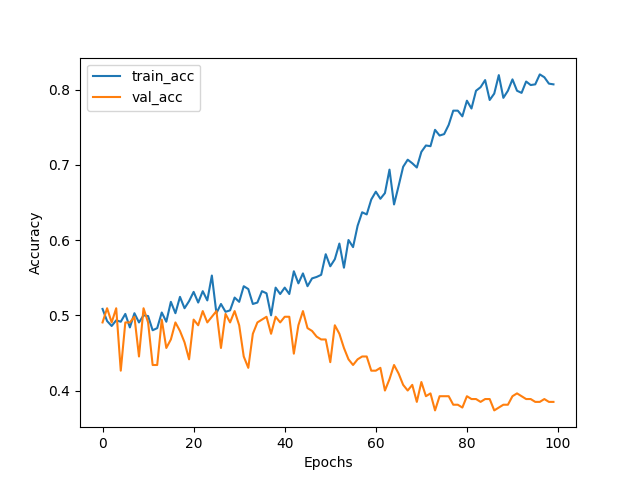}
		\caption{Exp. 1 ViT - FF++}
		\label{exp1.10}
	\end{minipage}
\end{figure}

With the exception of the DFTimit LQ and UADFV datasets, the model shows clear overfitting from episodes 50/60 onwards. This is easily noticeable in the learning curves of figures \ref{exp1.1} to  \ref{exp1.10}.

Considering that the model architecture is distributed as follows:

\begin{enumerate}
    \item ViT {encoder}
    \item Latent space
    \item Classifier
\end{enumerate}

The overfitting obtained in experiment 1 could occur in both the {encoder} and the classifier. Therefore, it was decided to introduce \textit{dropout}, a technique that reduces overfitting by randomly omitting selected nodes.

A \textit{dropout} with a probability of 0.2 was added to the multi-layer perceptron used as a classifier.

Additionally, a \textit{scheduler} was added so the learning rate can be adapted to the gradient descent optimization procedure, thus, improving performance and reducing training time.

There are various types, but it was decided to use the CosineAnnealingLR \cite{cosine}, which adjusts the learning rate according to the cosine journey, combining periods with higher and lower values. This helps avoid model stagnation in local minima and converges to points closer to the optimal global minimum.

The configuration for the model was as follows:

latex
\begin{itemize}
    \item {Initial learning rate}: 0.001 with CosineAnnealingLR {scheduler}.
    \item {Episodes}: 100.
    \item {{Batch Size}}: 32.
    \item {Transformer Dimension}: 192.
    \item {Transformer Depth}: 4.
    \item {Transformer Heads}: 3.
    \item {Transformer Head Dimension}: 64.
    \item {{Dropout} probability in classifier}: 0.2.
\end{itemize}

\begin{figure}[ht!]
	\centering
	\begin{minipage}{.45\columnwidth}
		\centering
		\includegraphics[width=\textwidth]{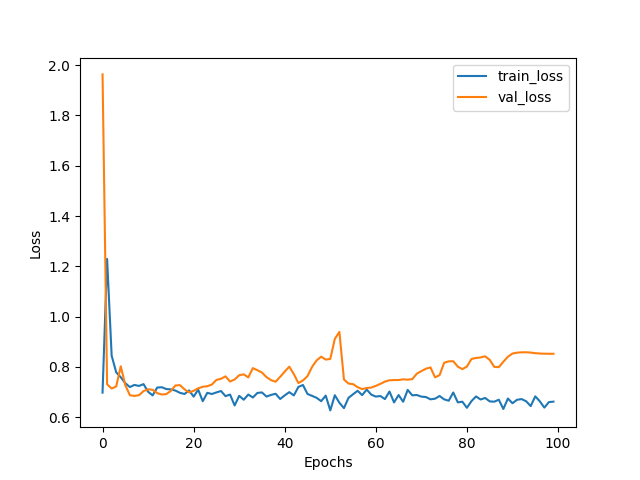}
		\caption{Exp. 2 ViT - UADFV}
		\label{exp2.1}
	\end{minipage}%
	\begin{minipage}{.45\columnwidth}
		\centering
		\includegraphics[width=\textwidth]{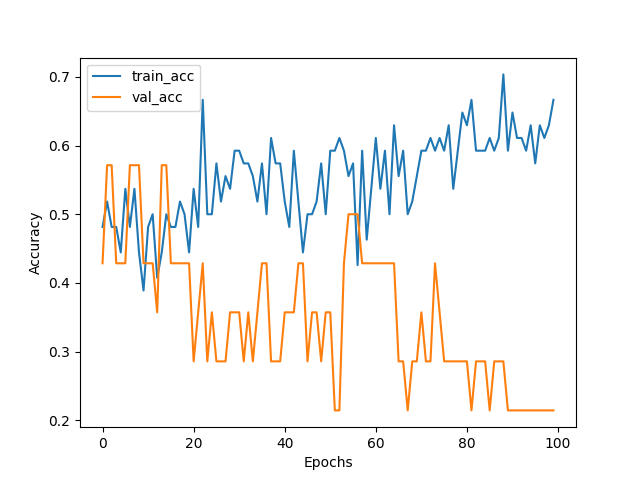}
		\caption{Exp. 2 ViT - UADFV}
		\label{exp2.2}
	\end{minipage}
 	\begin{minipage}{.45\columnwidth}
		\centering
		\includegraphics[width=\textwidth]{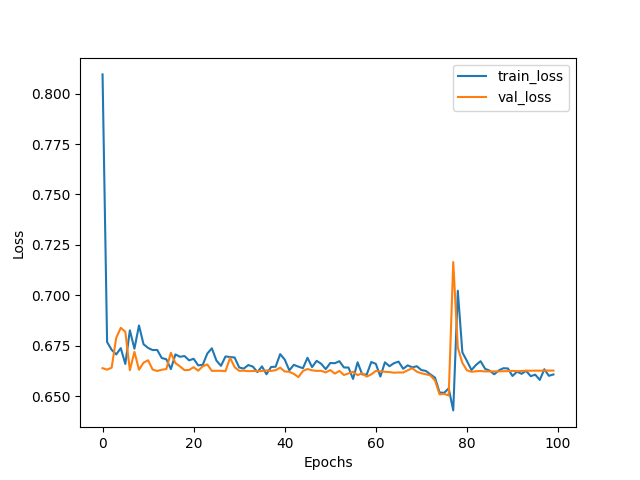}
		\caption{Exp. 2 ViT - DFTimit LQ}
		\label{exp2.3}
	\end{minipage}
 	\begin{minipage}{.45\columnwidth}
		\centering
		\includegraphics[width=\textwidth]{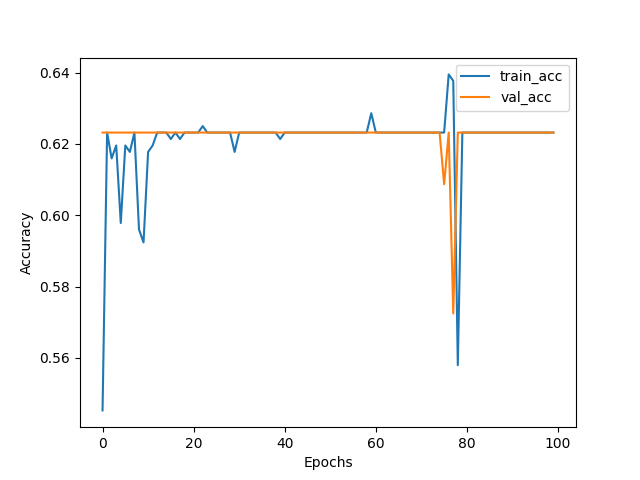}
		\caption{Exp. 2 ViT - DFTimit LQ}
		\label{exp2.4}
	\end{minipage}
 	\begin{minipage}{.45\columnwidth}
		\centering
		\includegraphics[width=\textwidth]{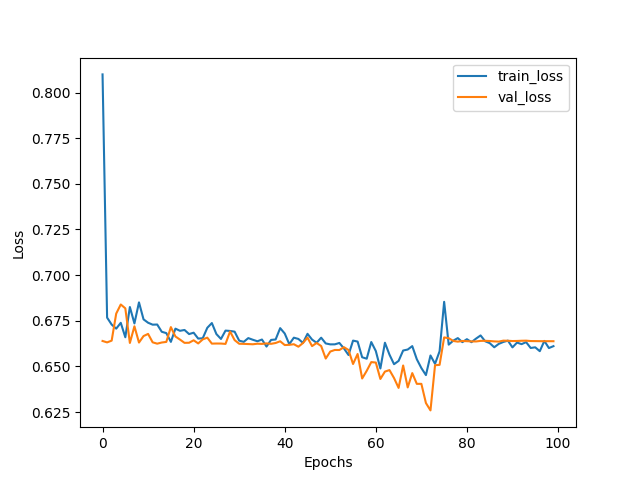}
		\caption{Exp. 2 ViT - DFTimit HQ}
		\label{exp2.5}
	\end{minipage}
 	\begin{minipage}{.45\columnwidth}
		\centering
		\includegraphics[width=\textwidth]{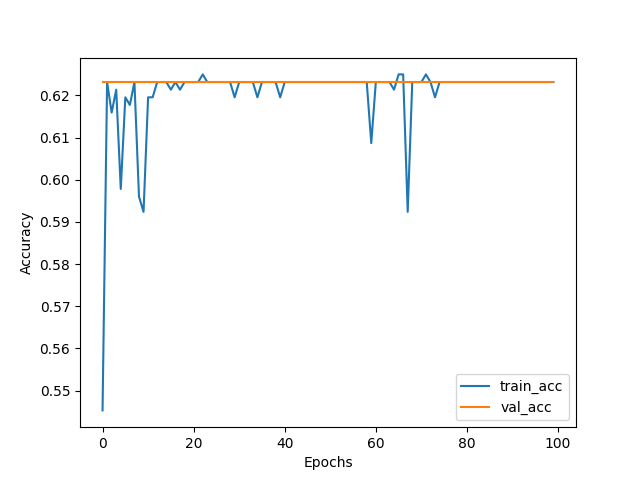}
		\caption{Exp. 2 ViT - DFTimit HQ}
		\label{exp2.6}
	\end{minipage}
  	\begin{minipage}{.45\columnwidth}
		\centering
		\includegraphics[width=\textwidth]{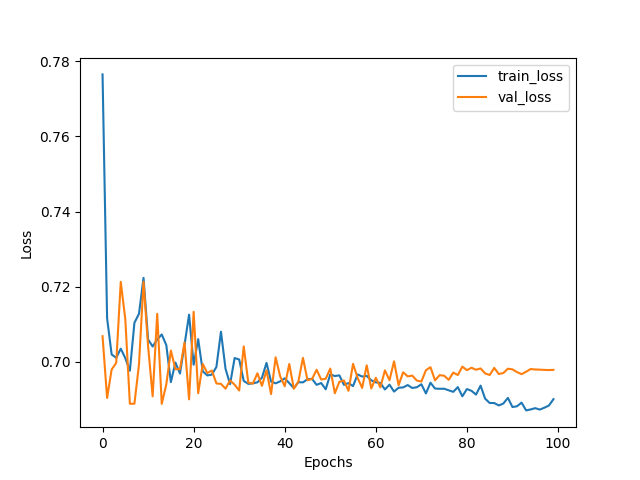}
		\caption{Exp. 2 ViT - DFDC}
		\label{exp2.7}
	\end{minipage}
 	\begin{minipage}{.45\columnwidth}
		\centering
		\includegraphics[width=\textwidth]{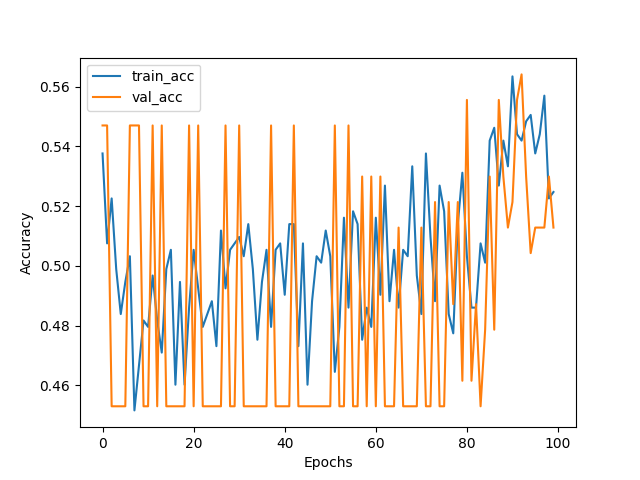}
		\caption{Exp. 2 ViT - DFDC}
		\label{exp2.8}
	\end{minipage}
 	\begin{minipage}{.45\columnwidth}
		\centering
		\includegraphics[width=\textwidth]{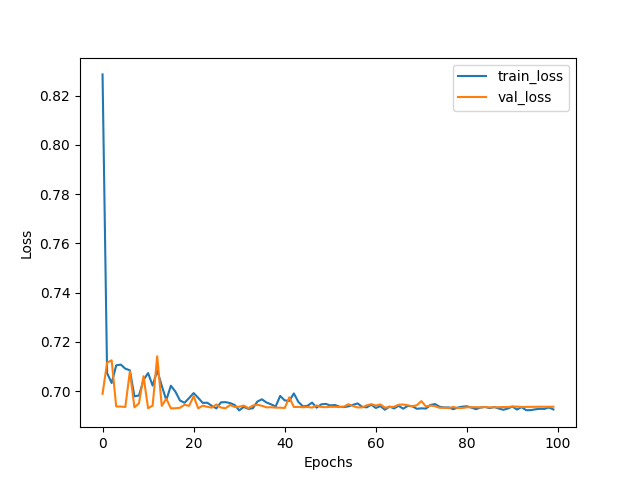}
		\caption{Exp. 2 ViT - FF++}
		\label{exp2.9}
	\end{minipage}
 	\begin{minipage}{.45\columnwidth}
		\centering
		\includegraphics[width=\textwidth]{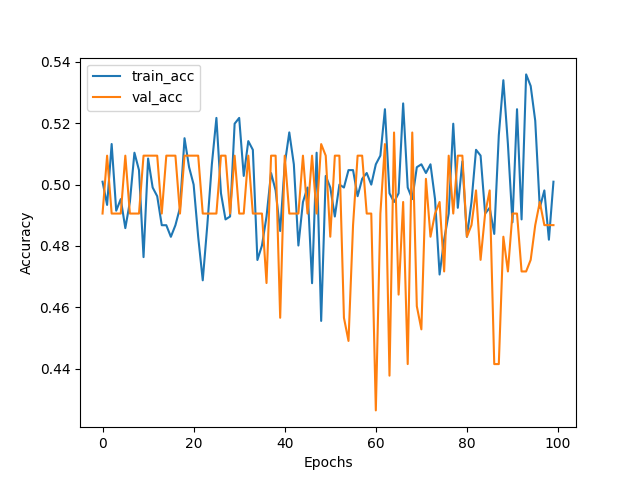}
		\caption{Exp. 2 ViT - FF++}
		\label{exp2.10}
	\end{minipage}
\end{figure}

Figures \ref{exp2.1} to \ref{exp2.10} show the learning curves of the experiment with different datasets. The graphs reflect that the overfitting problem has been resolved with the introduction of \textit{dropout}, confirming that the issue was in the classifier.

However, these curves do not show a considerable increase in learning with graphs, sometimes exhibiting abrupt changes. This may be due to a \textit{batch size}, learning rate, or model dimensions lower than necessary.

Considering the results, it was decided to increase the dimensionality of the transformer representations from 192 to 512 and the number of heads from 3 to 8. This decision was based on the notion that larger model sizes generally yield better results, as suggested by the state of the art \cite{attention}. As a consequence derived from the resource constraints, the number of episodes was reduced from 100 to 50 due to the increased time required with the augmented model dimensionality.

\begin{table}
  \centering
\begin{tabular}{|c|c|c|}
\hline
\rowcolor[HTML]{C0C0C0} 
DataSet & Precision & Loss function \\ \hline
\rowcolor[HTML]{FFFFFF} 
{\color[HTML]{1F2328} UADFV} & {\color[HTML]{1F2328} 0.5714} & {\color[HTML]{1F2328} 0.6835} \\ \hline
\rowcolor[HTML]{EFEFEF} 
DFTimit LQ & {\color[HTML]{1F2328} 0.6250} & {\color[HTML]{1F2328} 0.6769} \\ \hline
\rowcolor[HTML]{FFFFFF} 
{\color[HTML]{1F2328} DFTimit HQ} & {\color[HTML]{1F2328} 0.6756} & {\color[HTML]{1F2328} 0.6936} \\ \hline
\rowcolor[HTML]{EFEFEF} 
{\color[HTML]{1F2328} FaceForensics++} & {\color[HTML]{1F2328} 0.5094} & {\color[HTML]{1F2328} 0.6930} \\ \hline
DFDC & \cellcolor[HTML]{FFFFFF}{\color[HTML]{1F2328} 0.5470} & 0.6888 \\ \hline
\end{tabular}
\caption{Results \textit{Vision Transformer}, learning rate 0.001}
\label{vit-exp3}
\end{table}

As shown in Table \ref{vit-exp3}, presenting the results of this experiment, there has been a generic improvement in performance by increasing the model dimensions, reaching up to 67.56\% precision on the DFTimit HQ dataset.

Since it is not possible to further increase the model dimensions due to the available computational capabilities, it was decided to conduct a final experiment by increasing the learning rate to see if it could improve the results, considering that a too-small rate might have been used for the model. Thus, the same model as in Experiment 3 was used, with an initial learning rate increased to 0.01, and the number of episodes reduced from 50 to 30.

The results show a slight improvement in almost all datasets compared to Experiment 3. This indicates that the learning rate used in the previous experiment was too conservative, and it could be increased, also reducing the execution time of the experiments.

The results of this experiment, presented in Table \ref{vit-exp4}, show better outcomes for the UADFV, DFTimit HQ, and FaceForensics++ datasets, increasing to 62.31\%, 62.31\%, and 55.0\%, respectively. On the other hand, DFDC maintains the same precision, and the DFTimit LQ dataset decreases its effectiveness from 62.5\% to 57.14\%. Therefore, it can be observed that depending on the data to which this learning rate modification is applied, it may be more or less effective for the model's performance.

Unfortunately, even with the proposed parameter fine-tuning, results are insufficient to fully automate the deepfake detection process. However it can help to detect a reduced proportion of deepfakes when used as a first filter for manual interventions (e.g. for fact-checkers).

\begin{table}
\centering
\begin{tabular}{|c|c|c|}
\hline
\rowcolor[HTML]{C0C0C0} 
\textbf{DataSet} & \textbf{Precision} & \textbf{Loss function} \\ \hline
\rowcolor[HTML]{FFFFFF} 
{\color[HTML]{1F2328} UADFV} & {\color[HTML]{1F2328} 0.6231} & {\color[HTML]{1F2328} 0.6668} \\ \hline
\rowcolor[HTML]{EFEFEF} 
DFTimit LQ & {\color[HTML]{1F2328} 0.5714} & {\color[HTML]{1F2328} 0.6829} \\ \hline
\rowcolor[HTML]{FFFFFF} 
{\color[HTML]{1F2328} DFTimit HQ} & {\color[HTML]{1F2328} 0.6231} & {\color[HTML]{1F2328} 0.6626} \\ \hline
\rowcolor[HTML]{EFEFEF} 
{\color[HTML]{1F2328} FaceForensics++} & {\color[HTML]{1F2328} 0.5501} & {\color[HTML]{1F2328} 0.7128} \\ \hline
DFDC & \cellcolor[HTML]{FFFFFF}{\color[HTML]{1F2328} 0.5470} & 0.6891 \\ \hline
\end{tabular}
\caption{Results Vision Transformer, learning rate 0.01}
\label{vit-exp4}
\end{table}

\section{Conclusions}

This work aimed to explore different artificial intelligence models to detect deepfakes in various datasets and assess their performance in an environment with limited computational capabilities.

The initial experiments were conducted using a three-dimensional convolutional neural network (3DCNN), a type of network commonly employed in models with multimedia input data. Specifically, a (2+1)D model with ResNet was chosen, which had shown superior results in various state-of-the-art studies due to its separation of spatial and temporal components. However, the high computational and hardware requirements posed significant limitations. After the first two experiments yielded results only slightly better than chance, the model was discarded for this scenario.

The alternative model employed in this study is a Vision Transformer (ViT), typically used for image classification. Design decisions were taken to cope with the existing limitations. Results were gradually improved through various experiments, addressing issues such as overfitting of the multi-layer perceptron classifier and fine-tuning model parameters. The achieved precision reached up to 67.56\% in some datasets. Limited computational capabilities hinder further model improvement, given the requirements of deep learning models and constraints on training. This constrains the feasibility of these models. However, in contrast to the previous case, there might be some applications for this or similar models as tools supporting manual interventions for stakeholders with limited resource capacity.

In summary, as anticipated, certain models are entirely impractical in these scenarios, while others might yield some results if prudent decisions are made, as shown with the ViViT. The selection of model hyperparameters, including dimensionality and learning rate, plays a pivotal role in influencing the model's performance. Achieving a delicate balance between model size and computational resources is crucial. The effects of adjusting hyperparameters differ across diverse datasets, indicating that a universal approach may not be optimal.

Future work will concentrate on validating the performance of alternative techniques under constrained computational capabilities, including approaches that leverage biometric indicators.

\begin{acknowledgements}
This work is part of the R\&D project PID2022-136684OB-C21 funded by the Spanish Ministry of Science and Innovation MCIN/AEI/10.13039/501100011033.
\end{acknowledgements}

\begingroup
\titleformat{\subsection}[runin]
  {\normalfont\bfseries\small}{\thesubsection}{17pt}{}
\subsection*{Data availability} 
The data that supports this work are openly available in FaceForensics\cite{fforensics} at \url{https://github.com/ondyari/FaceForensics}, UADFV\cite{UADFV} at \url{https://github.com/danmohaha/WIFS2018\_In\_Ictu\_Oculi} (filling a form), DeepfakeTIMIT\cite{TIMIT}  at \url{https://www.idiap.ch/en/dataset/deepfaketimit} and \url{https://conradsanderson.id.au/vidtimit/}, and DFDC\cite{dfdc} at \url{https://www.kaggle.com/c/deepfake-detection-challenge/data} and through \url{https://ai.meta.com/datasets/dfdc/}  

\subsection*{Conflict of interest} The authors declare that they have no known competing financial interests or personal relationships that could have appeared to influence the work reported in this paper.

\subsection*{Ethical approval} This article does not contain any studies with human participants or animals performed by any of the authors.
\endgroup
\bibliographystyle{abbrvnat}      
\bibliography{deepfake}

\end{document}